\documentclass[runningheads]{llncs}
\usepackage{graphicx}
\usepackage{tabularx}
\usepackage{booktabs}
\usepackage{amsmath}
\usepackage{subcaption}
\usepackage{algorithm}
\usepackage{algpseudocode}
\usepackage{multirow}
\usepackage{xcolor}

\makeatletter
\newcommand{\algmargin}{\the\ALG@thistlm}
\makeatother
\newlength{\whilewidth}
\algdef{SE}[parWHILE]{parWhile}{EndparWhile}[1]
  {\parbox[t]{\dimexpr\linewidth-\algmargin}{%
     \hangindent\whilewidth\strut\algorithmicwhile\ #1\ \algorithmicdo\strut}}{\algorithmicend\ \algorithmicwhile}%
\algnewcommand{\parState}[1]{\State%
  \parbox[t]{\dimexpr\linewidth-\algmargin}{\strut #1\strut}}

\newcolumntype{Y}{>{\centering\arraybackslash}X}

\graphicspath{ {images/} }

\newtheorem{hyp}{Hypothesis}

\makeatletter
\def\@opargbegintheorem#1#2#3{\trivlist
   \item[]{\bfseries #1\ #2\ (#3)} \itshape}
\makeatother

\begin{document}
\title{Augmenting Novelty Search with\\ a Surrogate Model to Engineer Meta-Diversity \\in Ensembles of Classifiers}
\titlerunning{Surrogate Model to Engineer Meta-Diversity}
\author{Rui P. Cardoso\inst{1} \and
Emma Hart\inst{4} \and
David Burth Kurka\inst{2} \and
Jeremy Pitt\inst{3}}
% \author{Anonymous 1\inst{1} \and Anonymous 2\inst{2}}
%
\authorrunning{Cardoso  et al.}
% \authorrunning{Anonymous et al.}
\institute{Imperial College London,
\email{rui.cardoso@imperial.ac.uk \and d.kurka@imperial.ac.uk \and j.pitt@imperial.ac.uk} \and
Edinburgh Napier University, 
\email{e.hart@napier.ac.uk}}
% \institute{University of Somewhere,
% \email{anonymous@somewhere.org}
% \and
% University of Place,
% \email{anonymous@place.org}}
%
\maketitle              % typeset the header of the contribution

\begin{abstract}
Using Neuroevolution combined with Novelty Search to promote behavioural diversity is capable of constructing high-performing ensembles for classification. However, using gradient descent to train evolved architectures during the search can be computationally prohibitive. Here we propose a method to overcome this limitation by using a \emph{surrogate} model which estimates the behavioural distance between two neural network architectures required to calculate the sparseness term in Novelty Search. We demonstrate a speedup of 10 times over previous work and significantly improve on previous reported results on three benchmark datasets from Computer Vision --- CIFAR-10, CIFAR-100, and SVHN. This results from the expanded architecture search space facilitated by using a surrogate. Our method represents an improved paradigm for implementing horizontal scaling of learning algorithms by making an explicit search for diversity considerably more tractable \emph{for the same bounded resources}. Paper accepted at EvoStar 2022.

\end{abstract}

\section{Introduction}\label{sec:intro}

Ensemble performance is fundamentally dependent on both the accuracy of individual base learners and the diversity between them \cite{dietterich2000ensemble}. However, techniques to promote diversity are typically only implicit, such as training the models on different subsets of the data or starting from different random initialisations. In previous work \cite{10.1145/3449639.3459308}, we proposed a method that \emph{explicitly} searched for diversity amongst a set of base learners by making use of metrics for measuring \emph{behavioural} diversity. However, a fundamental limitation of this approach was its computational complexity, with a costly step of training all the neural network models in the population at each step of the search. Such time and computational demands compromise the goal of our approach, which is to develop learning algorithms which scale horizontally, namely with models which can be distributed across many low-cost machines. The claims to tackling the unwieldiness of Deep Learning (DL) algorithms with more scalable solutions were undermined by the fact that our approach was so computationally intensive.

In order to overcome the costly step of training each model, we have introduced a surrogate model \cite{Siems2020} into our method. We use this surrogate model, pretrained on a sample drawn from the search space of neural network architectures, to get an estimate of the \emph{error} distance between two neural networks given architectural descriptors, \emph{without} training these networks. Whereas this calculation had previously been a very costly step, this technique renders it essentially instantaneous. This produces a speedup of 10 times compared to the previous approach when the same parameters are used, \emph{without loss of performance}. By changing the parameters to expand the search space of neural network architectures we have considerably improved on previous results reported on three benchmark datasets from Computer Vision --- CIFAR-10, CIFAR-100, and SVHN. Using a surrogate model has enabled us to search a wider space of neural network architectures and run the Novelty Search procedure for longer.

The major contribution of this paper is that it proposes an improved paradigm for implementing horizontal scaling of learning algorithms. Explicitly creating diversity amongst the members of an ensemble establishes a sound criterion for distributing these models. By improving the method with a surrogate model in the way described above, our approach makes an explicit search for diversity considerably more tractable \emph{for the same bounded resources}.

\section{Background}\label{sec:background}
% general surrogate models with neuro-e

In previous work \cite{10.1145/3449639.3459308}, we proposed a method that \emph{explicitly} searched for behavioural diversity amongst a set of base learners. This used a Novelty Search (NS) algorithm in conjunction with Neuroevolution, in which novelty was determined by novel metrics that explicitly measured behavioural diversity. The evolved behaviourally diverse ensembles outperformed both their individual learners and ensembles created with techniques that only implicitly promote diversity. This work also enabled us to study and compare different definitions of diversity. However, a fundamental limitation was its computational complexity. In order to calculate the behavioural distance between two models, we need to compare the classification errors that they make on a validation data set. This requires first training each member of the current population of neural network models on a training data set with gradient descent at each iteration of the NS. If computational resources are limited, this very time-consuming step can be prohibitive. This poses significant challenges because it restricts the search to only a few iterations at best and renders the problem intractable at worst. Here we overcome this difficulty by augmenting the NS with a \emph{surrogate model}.

Combining an evolutionary algorithm (EA) with a surrogate modelling function has been common in the literature for many years, e.g. in single-objective optimisation \cite{tong2021surrogate}, multi-objective optimisation \cite{ruan2020surrogate}, and particularly in expensive optimisation \cite{zhou2006combining}. A first surrogate model for neural network optimisation was introduced by Gaier {\em et al.} \cite{gaier2018data} and used in conjunction with the NEAT \cite{stanley2002evolving} algorithm for evolving the weights and topology of a neural network. This paper used a surrogate distance-based model, employing a genotypic compatibility distance metric that is part of NEAT. The approach has been quickly adopted in the literature using a range of surrogates and a variety of methods to evolve networks. There are several examples of approaches that use surrogates to estimate the performance of an architecture. For example, in 2017 Deng \emph{et al.} proposed the Peephole algorithm \cite{deng2017peephole}, which predicted the performance of a convolutional neural network based on its architecture information: a long-short term memory (LSTM) neural network was used to train the model. Stork \emph{et al.} \cite{stork2019improving} extended a Cartesian Genetic Programming method called CPGANN to evolve neural networks using surrogate-based optimisation to reduce the number of fitness evaluations required. They used a Kriging model \cite{kriging} as the surrogate. In \cite{sun2019surrogate}, a Random Forest algorithm (RF) was used as a surrogate to predict the performance of a CNN architecture --- the authors proposed a method for describing a CNN as a set of features which were used as input to the RF. In \cite{Siems2020}, the authors use a surrogate benchmark for neural architecture search (NAS).

In contrast, Hagg \textit{et al.} \cite{Hagg2019} introduce a more flexible method for building a surrogate model that is independent of network topology: rather than describing the neural network architecture, they introduce a \textit{phenotypic} metric which measures the difference in output between two neural networks given the same input sequence. The difference is used in a Kriging surrogate model. Our proposed approach is conceptually closest to that of Hagg. For a given neural network, we calculate a behavioural vector that describes its behaviour on a dataset (see Section \ref{sec:metrics}). We then propose a RF surrogate model that is used to estimate the distance between the behavioural vectors produced by any two neural networks, as this value is required to drive a NS algorithm.

\section{Methods and Materials}\label{sec:methods}

We use NS to evolve an ensemble of behaviourally diverse neural network models. The NS operates over a space of architectures defined by a set of hyperparameters. Unlike our previous work \cite{10.1145/3449639.3459308}, where the neural networks in a generation had to be trained with gradient descent at each iteration of the NS in order to calculate the behavioural distances between each pair of architectures, these distances are now \emph{estimated} by a surrogate model which is pretrained on a sample drawn from the space of neural network architectures. The most diverse models are added to the final ensemble, which is then trained on the input data. We evaluate the method against our previous method and compare the performance obtained with different diversity metrics. The following subsections go into detail about each of these steps.

\subsection{Neural Network Architectures}\label{sec:arch}

\begin{figure}
\begin{subfigure}{.45\textwidth}
  \centering
  \includegraphics[scale=.15]{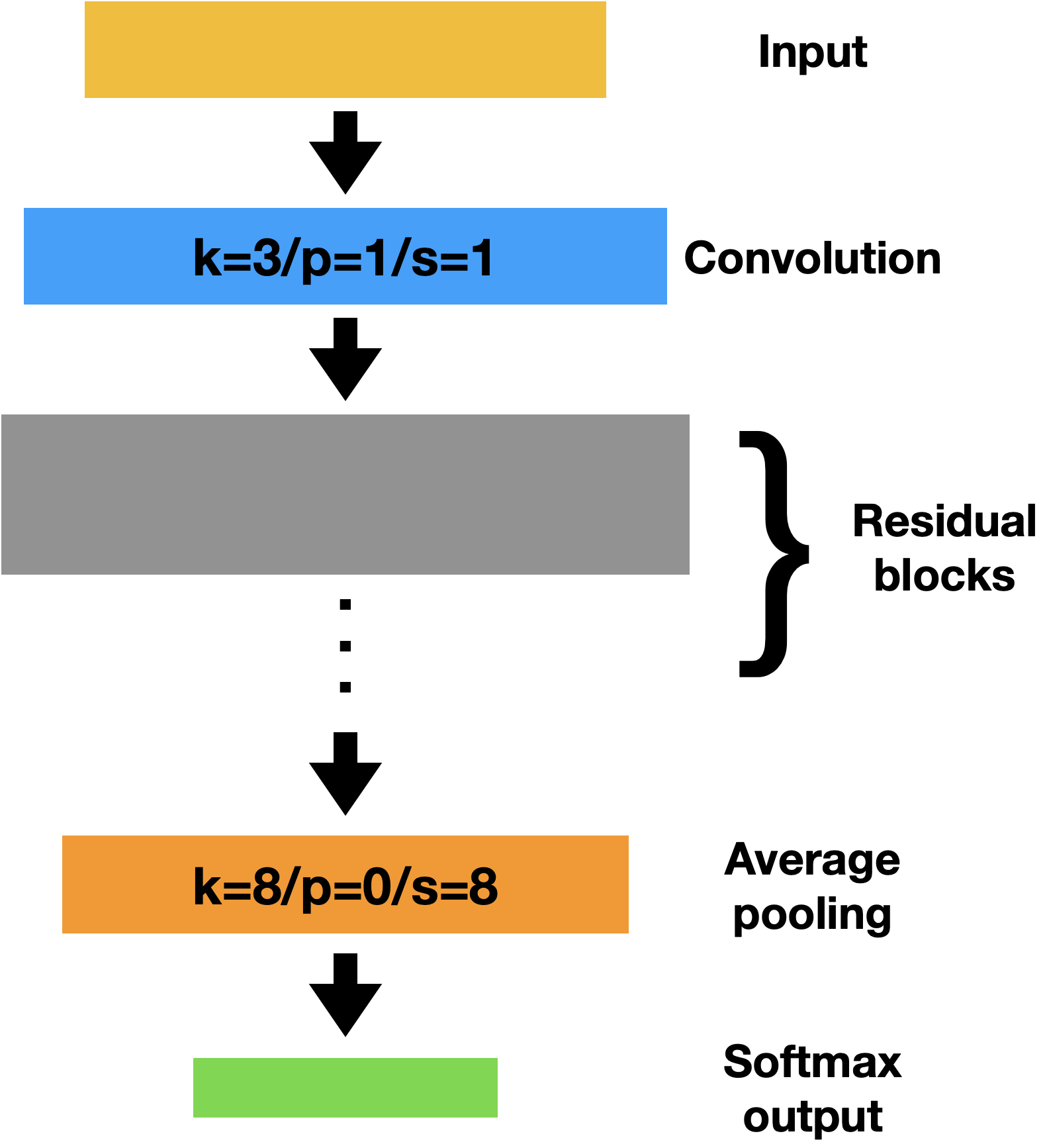}
    \caption{Generic residual neural network (k = kernel size; p = padding; s = stride)}
    \label{fig:network}
\end{subfigure}
\hfill
\begin{subfigure}{.45\textwidth}
  \centering
  \includegraphics[scale=.15]{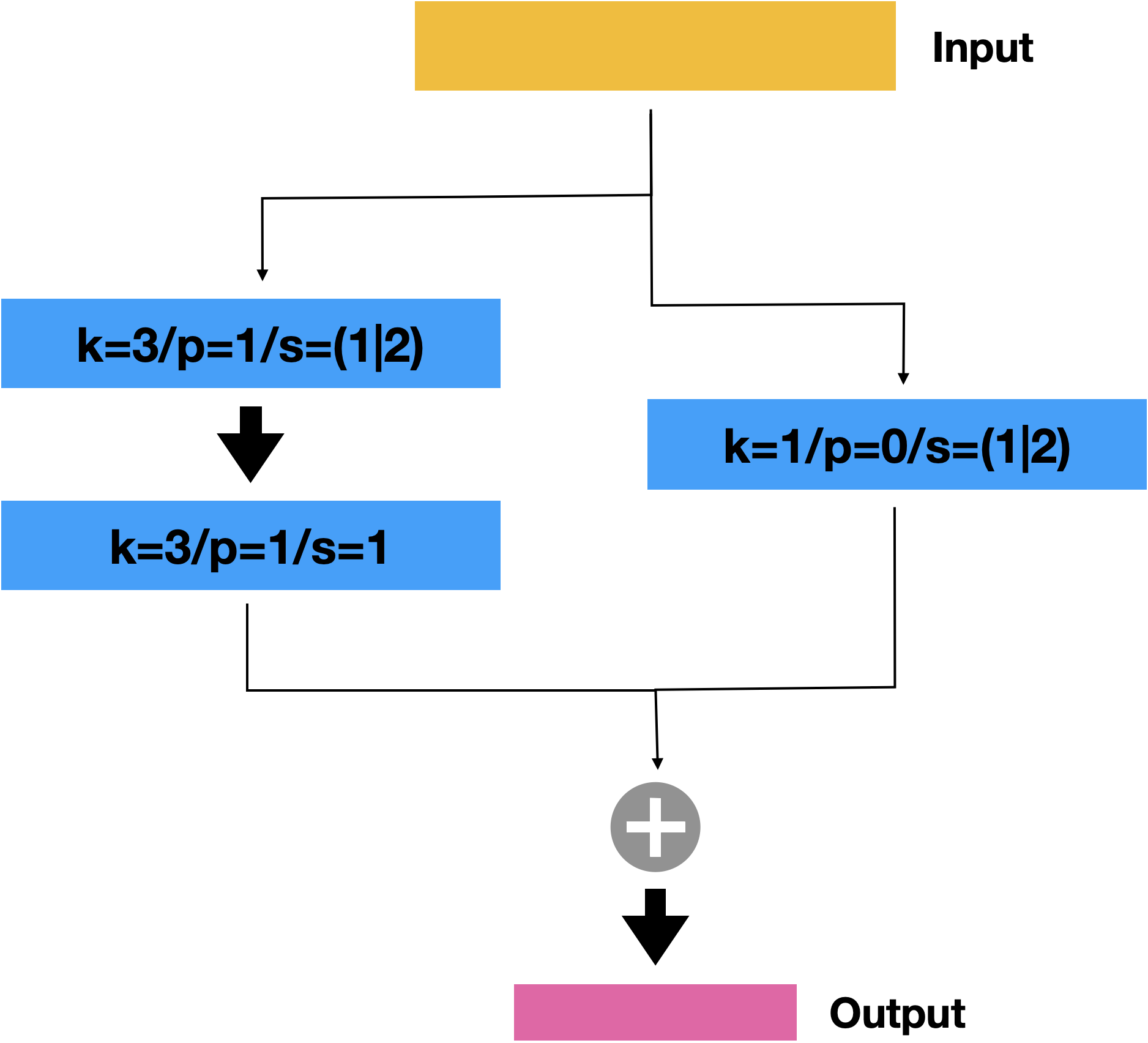}
  \caption{Generic residual block}
  \label{fig:res_block}
\end{subfigure}
\caption{Generic topology of individual neural networks}
% \vspace{-20pt} 
\label{fig:topology}
\end{figure}

The architectures evolved by our procedure are \emph{residual neural networks} \cite{he2016deep} based on the wide architectures proposed by \cite{zagoruyko2016wide}. They are of the same kind as those we used in our previous work. Figure \ref{fig:network} shows a generic neural network and Figure \ref{fig:res_block} illustrates a generic residual block. Please refer to our previous paper \cite{10.1145/3449639.3459308} for a more detailed description of these architectures.

The \emph{hyperparameters} of each network are evolved by NS. Each individual in the population is defined by a variable-length vector, depending on the number of blocks $r$: $[J, C, O^1, ..., O^r, D^1, ..., D^r]$, where $J$ is a Boolean value indicating whether the network should be trained jointly or separately if it is in the final ensemble, $C$ is the output size of the first convolution, $O^i$ is the output size of block $i$, and $D^i$ its dropout probability. Each individual is mapped to a Pytorch module \cite{Paszke2019} for implementation purposes. The \emph{parameters} of each network are randomly initialised and then optimised by a standard gradient descent procedure.

In order to preprocess the input to the surrogate model, we \emph{normalise} the representation described above in the following way: we first \emph{rescale} the elements in all positions so that they lie between 0 and 1. The first element is the Boolean value indicating whether the neural network should be trained jointly or separately, so it need not be normalised. Then, given that the representations have variable length depending on the number of residual blocks in each neural network, we \emph{pad} the vector so that it has fixed length, corresponding to the maximum possible number of residual blocks, by adding an appropriate number of elements equal to 0 before the sequence of block output sizes and before the sequence of dropout probabilities. Therefore, if the number of residual blocks in the network is $r$ and the maximum number of blocks is $R$; if the maximum and minimum sizes of the first convolution in the network are $C_{max}$ and $C_{min}$, respectively; if the maximum and minimum sizes of each residual block are $O_{max}$ and $O_{min}$, respectively; and if the maximum and minimum dropout probability of each block are $D_{max}$ and $D_{min}$; then the normalised representation of neural network $m_i$ is:

\begin{equation}\label{eq:norm}
    \begin{aligned}
        \textbf{norm\_rep}_i = & \left[ J_i, \frac{C_i - C_{min}}{C_{max} - C_{min}}, \right. \\
        & \left. 0, ..., 0, \frac{O_i^{R-r} - O_{min}}{O_{max} - O_{min}}, ..., \frac{O_i^{R} - O_{min}}{O_{max} - O_{min}}, \right. \\
        & \left. 0, ..., 0, \frac{D_i^{R-r} - D_{min}}{D_{max} - D_{min}}, ..., \frac{D_i^{R} - D_{min}}{D_{max} - D_{min}} \right]
    \end{aligned}
\end{equation}

Where there are $R-r$ elements equal to 0 before the sequence of block output sizes and before the sequence of dropout probabilities.

\subsection{Diversity Metrics}\label{sec:metrics}

In order to calculate novelty scores, which are used as the objective function by the NS, we have considered six different diversity metrics, five of which we have defined ourselves. These metrics are calculated between each pair of individual neural network architectures. We have used three of these metrics in the previous version of our procedure \cite{10.1145/3449639.3459308}.

Let $\boldsymbol{y}_i$ be the vector of predictions for model $m_i$ with each prediction $y_i^n$ for data point $\boldsymbol{x}^n$ being a class label in $\{1..C\}$. Let $\boldsymbol{p}_i$ be a binary vector where $p_i^n = 1$ if the prediction $y_i^n$ is correct and $p_i^n = 0$ otherwise. Let $N^{11}$, $N^{00}$, $N^{01}$, and $N^{10}$, respectively, be the total number of test instances where two models are both correct, both incorrect, and when one is correct and the other is not. The first diversity metric we consider is the \emph{proportion of different errors} between two models when at least one of them is \emph{correct}. We expect it to provide insight into the divergence between the errors made by two models. It is defined as:

\begin{equation}
    \text{prop}^1_{i,j} = \frac{N^{01} + N^{10}}{N^{11} + N^{01} + N^{10}}
\end{equation}

The second diversity metric we consider is very similar and is the \emph{proportion of different errors} between two models when at least one of them is \emph{incorrect}. We have first proposed this metric in our previous work \cite{10.1145/3449639.3459308}, defining it as:

\begin{equation}
    \text{prop}^2_{i,j} = \frac{N^{01} + N^{10}}{N^{00} + N^{01} + N^{10}}
\end{equation}

The third metric we propose is the \emph{harmonic mean} between these two proportion metrics. This is a sound way of averaging the two proportion metrics into a single metric so that they are both taken into account. It is defined as:

\begin{equation}
    \text{prop}^{\text{harm}}_{i,j} = \frac{2 \cdot \text{prop}^1_{i,j} \cdot \text{prop}^2_{i,j}}{\text{prop}^1_{i,j} + \text{prop}^2_{i,j}}
\end{equation}

We also consider a widely used metric (e.g. \cite{phdthesis,Kuncheva2003,Pasti2010625}) defined as the \emph{disagreement} between two models, i.e. the proportion of test instances where one is correct and the other is not. We take this metric into account since it expresses how commonly two models disagree on any test instance. It is defined as:

\begin{equation}
    \text{dis}_{i,j} = \frac{N^{01} + N^{10}}{N^{00} + N^{01} + N^{10} + N^{11}}
\end{equation}

Consider now the \emph{two's complement} of the binary vector of correct predictions $\boldsymbol{p}_i$, $\boldsymbol{w}_i$, i.e. the binary vector of \emph{wrong} predictions. The next metric we propose is the \emph{cosine distance} between the binary vectors of wrong predictions made by two models $m_i$ and $m_j$. Like $\text{prop}^1_{i,j}$ and $\text{prop}^2_{i,j}$, we consider this metric because it is a measure of the distance between the errors made by two models. We have defined it as:

\begin{equation}
    \text{cos\_dist}_{i,j} = 1 - \frac{\boldsymbol{w}_i \cdot \boldsymbol{w}_j}{\lVert \boldsymbol{w}_i \rVert \lVert \boldsymbol{w}_j \rVert}
\end{equation}

At last we consider a metric of \emph{architectural} diversity. Take the \emph{normalised} vector which represents each individual neural network, as described in section \ref{sec:arch}. Let its size be $L$. To obtain an architectural representation, we simply remove the first element from the normalised representation, i.e. the Boolean value indicating whether or not the neural network should be trained separately or jointly. Thus, referring to Equation \ref{eq:norm}, the architectural representation of model $m_i$ is:

\begin{equation}
    \textbf{arch\_rep}_i = \textbf{norm\_rep}_i^{\{1..L-1\}}
\end{equation}

We then define \emph{architectural distance} between neural networks $m_i$ and $m_j$ as the cosine distance between their normalised architectural representations:

\begin{equation}
    \text{arch\_dist}_{i,j} = 1 - \frac{\textbf{arch\_rep}_i \cdot \textbf{arch\_rep}_j}{\lVert \textbf{arch\_rep}_i \rVert \lVert \textbf{arch\_rep}_j \rVert}
\end{equation}

These metrics determine the \emph{behavioural distance} between two neural network models, which is used to calculate the novelty scores that guide the NS procedure, as explained in Section \ref{sec:ns}. Note that the metrics $\text{prop}^2_{i,j}$ and $\text{cos\_dist}_{i,j}$ focus more closely on the instances where the models made a \emph{prediction error}. In our previous work \cite{10.1145/3449639.3459308}, these two metrics have led to better performance than the others. Here we are interested in learning whether the same pattern can be observed with our new improved version of the NS procedure.

\subsection{Surrogate Model to Estimate Distances}\label{sec:surrogate}

The NS requires novelty scores to be determined, which in turn require the distances between pairs of neural networks in the current population to be calculated. However, calculating the exact distance values between two neural network models entails first training the models on the input data with gradient descent and then evaluating them on a validation dataset, as we did in previous work \cite{10.1145/3449639.3459308}. This can be a very costly step if computational resources are limited, which constrains the NS to only a few iterations and the population to a small size --- as the neural networks have to be trained in parallel for efficiency. Here we overcome this limitation by pretraining a Random Forest \cite{breiman2001random} surrogate model which estimates the behavioural distances between a pair of neural network models.

\begin{figure}
\begin{subfigure}[b]{.45\textwidth}
  \centering
  \includegraphics[scale=0.18]{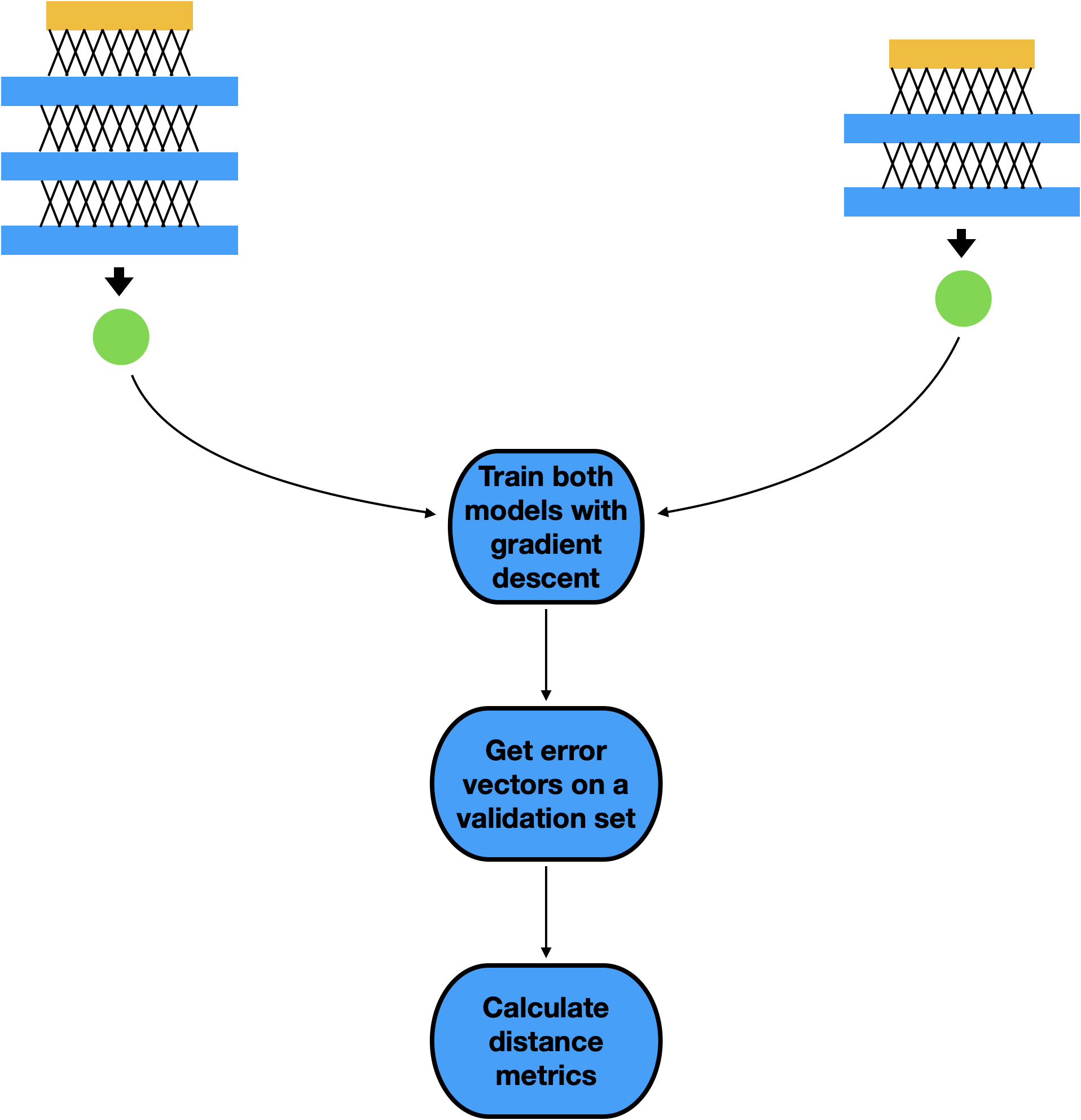}
    \caption{Steps required to calculate exact distance values between two neural networks}
    \label{fig:no_surrogate}
\end{subfigure}
\hfill
\begin{subfigure}[b]{.45\textwidth}
  \centering
  \includegraphics[scale=0.17]{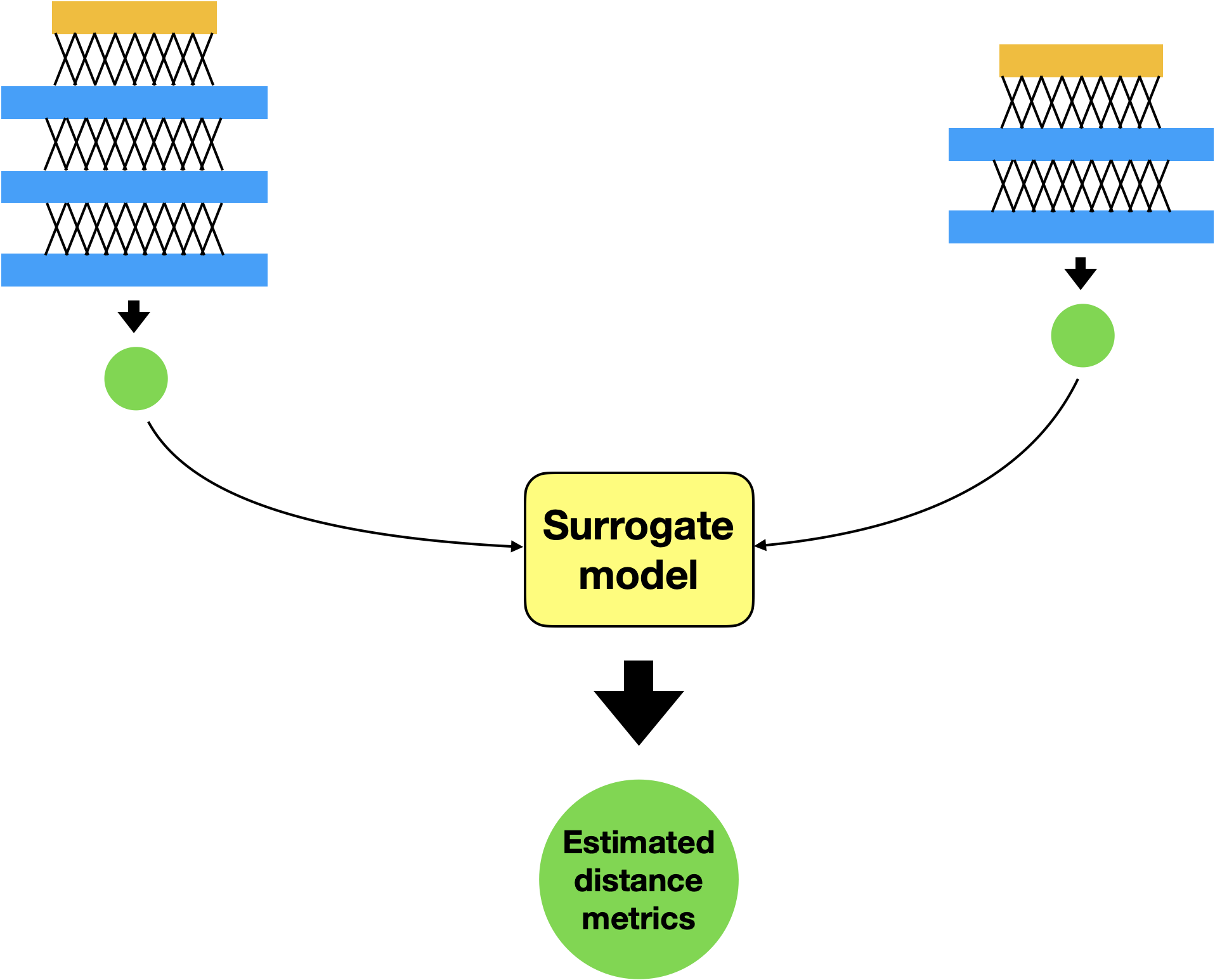}
  \caption{Estimating the distance values between two neural networks with a surrogate model}
  \label{fig:surrogate}
\end{subfigure}
\caption{Difference between calculating and estimating distance values}
\vspace{-30pt} 
\label{fig:diffs}
\end{figure}

Note that the estimates of behavioural distances produced by the surrogate model do not need to be very accurate. This is because, when calculating the novelty score of a particular individual neural network, we only need to know relative distances in order to determine nearest neighbours. This means that the surrogate model need only capture the general trends of growth of the distance values, even if the actual values are not very precise. This makes the use of a surrogate model very appropriate with no need for a very complex model. Figure \ref{fig:diffs} shows the differences between the previous method for calculating exact distance values, shown in Figure \ref{fig:no_surrogate}, and the current method using a surrogate model, shown in Figure \ref{fig:surrogate}. Calculating exact distance values is a very costly step, potentially requiring several GPU hours depending on the length of training. In contrast, estimating these distances by means of the surrogate model is an instantaneous process, once the surrogate model has been trained on sample data beforehand.

\subsection{Pretraining the Surrogate Model}

The surrogate model must be trained beforehand so that it can be used effectively during the NS to estimate the distance values between two neural network models. To do this, we draw a sample of neural networks from the search space of architectures defined by the set of hyperparameters used with the NS method. We first train each of these neural networks with gradient descent and calculate their error vectors on a validation data set. We then build random pairs of neural networks and calculate the exact distance values, for all six metrics considered, between them as a function of either their error vectors or their architectural descriptors, as explained in Section \ref{sec:metrics}. Finally, we construct a data set on which we fit a Random Forest regressor \cite{breiman2001random} which takes as input the normalised representations of two neural network architectures, as per Section \ref{sec:arch}, and has six outputs: the estimates of the distance values for all six metrics considered. We have selected a Random Forest model due to its low complexity and because we expect it to generalise well on new data, given that it is an ensemble model. Algorithm \ref{alg:surrogate} describes the process of training this surrogate model in pseudocode.

\begin{algorithm}[t]
    \caption{Pretraining the surrogate model on sample architectures}
    \label{alg:surrogate}
    \begin{algorithmic}
        \State draw a sample $S$ of neural networks from the search space defined by $J$, $\left[ C_{min}, C_{max} \right]$, $\left[ O_{min}, O_{max} \right]$, and $\left[ D_{min}, D_{max} \right]$;
        \State $\text{train}(S)$; \algorithmiccomment{Models trained jointly or separately according to the value of $J_i$}
        \For{neural network model $m_i \in S$}
            \State get error vector $e_i$ on validation set $\mathcal{D}_{val}$;
        \EndFor
        \State build $\frac{\lVert S \rVert ^2 - \lVert S \rVert}{2}$ unique pairs of neural networks;
        \State initialise dataset $\mathcal{D}_{dists} \gets \emptyset$;
        \For{each pair $m_i$, $m_j$}
            \State $\boldsymbol{d} \gets$ all 6 distance values; \algorithmiccomment{calculated as per Section \ref{sec:metrics}}
            \State $\text{norm\_rep}_i$, $\text{norm\_rep}_j$ are the normalised representations of $m_i$ and $m_j$;
            \State add data point $x \gets \left\{ \text{norm\_rep}_i, \text{norm\_rep}_j, \boldsymbol{d} \right\}$ to $\mathcal{D}_{dists}$;
        \EndFor
        \State train random forest model $rf$ on $\mathcal{D}_{dists}$;
        \State \Return random forest model $rf$;
    \end{algorithmic}
\end{algorithm}

% \begin{algorithm}[t]
%     \caption{Training a set of models (separately and jointly)}
%     \label{alg:training}
%     \begin{algorithmic}
%         \Procedure{train}{$S$}
%             \State $S_{sep} \gets \left\{ m_i \vert m_i \in S \land \neg J_i \right\}$
%             \State $S_{joint} \gets \left\{ m_i \vert m_i \in S \land J_i \right\}$
%             \For{neural network model $m_i \in S_{sep}$}
%                 \State train $m_i$ with gradient descent on training set \algorithmiccomment{separate loss}
%             \EndFor
%             \State train $S_{joint}$ with joint loss function \algorithmiccomment{calculated by first averaging Softmax probabilities}
%         \EndProcedure
%     \end{algorithmic}
% \end{algorithm}

\subsection{Novelty Search Algorithm}\label{sec:ns}

Our algorithm for building an ensemble implements NS as described by \cite{Lehman2011}, applying it to our problem domain. Algorithm \ref{alg:ns} presents the pseudocode for this procedure. The original training data is split into two sets, one for training and one for validation. The training set is used to train the final ensemble; it is also used to train the sample of neural network architectures drawn from the search space that is in turn used to pretrain the surrogate model. Whereas training each of these sample neural networks makes use of the entire training set, pretraining the surrogate model only requires the validation set, which is used to calculate exact distance values between pairs of neural networks.

Selection in NS is driven by the novelty score, which computes the sparseness at any point in the behavioural space. This sparseness is defined by one of the distance metrics of Section \ref{sec:metrics}. Areas with denser clusters of visited points are considered less novel and therefore rewarded less. This is defined as the average distance to the $K$-nearest neighbours of a point, calculated with respect to the other individuals in the current generation and to a stored \emph{archive} of previously sampled solutions. Hence, the novelty score is calculated as:

\begin{equation}\label{eq:ns}
    NS_i = \frac{1}{k}\sum_{k=0}^{K} \text{div\_metric}(m_i, \mu_k)
\end{equation}

Where  $\mu_k$ is the $k$th-nearest neighbour of $m_i$ with respect to the diversity metric $\text{div\_metric}_{i,j}$, selected from the metrics defined in Section \ref{sec:metrics}.

Individuals are selected for reproduction on the basis of their novelty scores using a tournament selection procedure. In the interests of promoting divergence and avoiding convergence, reproduction only uses mutation. Mutation either adds or removes a randomly chosen residual block from an individual, modifying input/output sizes at the mutation point as necessary; changes the output size and dropout probability of a random block; or swaps two consecutive blocks chosen at random.

After evaluating the entire population, $n_A$ randomly chosen individuals are added to the \textit{archive}, following the method suggested in \cite{Gomes2015}. In addition, the individual from the population with the highest \emph{elite score}, calculated in a similar fashion to the novelty score, is added to an \emph{elite archive}. After running the NS for the specified number of iterations, a subset of this elite archive is selected as the final ensemble. This subset is chosen so as to maximise the average distance amongst its members. The final ensemble is then trained by gradient descent, the \emph{only time} when this parameter optimisation takes place.

\begin{algorithm}[t]
    \scriptsize
    \caption{Ensemble evolution through NS}
    \label{alg:ns}
    \begin{algorithmic}
        \State randomly initialise population $pop$;
        \State $archive \gets \emptyset$;
        \State $elite\_archive \gets \emptyset$;
        \State draw $\mathcal{D}_{train}$ and $\mathcal{D}_{val}$ from training set $\mathcal{D}$;
        \State set evolution iterations $epochs$;
        \State set archive sample size $n_A$;
        \State set final ensemble size $ensemble\_size$;
        \State surrogate model $s_{div}$ pretrained as per Section \ref{sec:surrogate};
        \State select diversity $\text{div\_metric}_{i,j}$ from Section \ref{sec:metrics};
        \For{$epochs$}
            \For{$m_i, m_j \in pop \times pop \cup archive: m_i \neq m_j$}
                \State $\text{div\_metric}_{i,j} \approx s_{div}(m_i, m_j)$
            \EndFor
            \For{$m_i, m_j \in pop \times elite\_archive$}
                \State $\text{div\_metric}'_{i,j} \approx s_{div}(m_i, m_j)$
            \EndFor
            \For{$m_i \in pop$}
                \State $NS_i \gets \frac{1}{k}\sum_{k=0}^{K} \text{div\_metric}(m_i, \mu_k)$ \algorithmiccomment{Equation \ref{eq:ns}}
                \State $NS_i' \gets \sum_{m_j \in elite\_archive} \text{div\_metric}'(m_i, m_j)$ 
            \EndFor
            \State $sample \gets \text{random\_sample}(pop, n_A)$
            \State $archive \gets archive \cup sample$
            \State $el\_best \gets \text{max}(pop, NS_i')$
            \State $elite\_archive \gets elite\_archive \cup \{el\_best\}$
            \State $s \gets \text{tournament\_select}(pop, NS_i)$
            \State $pop \gets \text{mutate}(s)$
        \EndFor
        \For{$m_i, m_j \in elite\_archive \times elite\_archive: m_i \neq m_j$}
            \State $\text{div\_metric}_{i,j}^{\ast} \approx s_{div}(m_i, m_j)$
        \EndFor
        \For{$m_i \in elite\_archive$}
            \State $NS_i^{\ast} \gets \sum_{m_j \in elite\_archive: m_i \neq m_j} \text{div\_metric}^{\ast}(m_i, m_j)$ 
        \EndFor
        \State $ensemble \gets \text{max}(elite\_archive, NS_i^{\ast}, ensemble\_size)$
        \State $\text{train}(ensemble)$; \algorithmiccomment{Models trained jointly or separately according to the value of $J_i$}
    \end{algorithmic}
\end{algorithm}

\subsection{Evaluation of Evolved Ensembles}\label{sec:eval}

In order to evaluate the performance of the evolved ensemble, we use the stacking technique \cite{Wolpert1992}, which trains a linear model to weight the predictions of each individual learner. This linear model is trained for a configurable number of iterations on the validation set mentioned in Section \ref{sec:ns}. This is to avoid overfitting the test set.

\subsection{Baseline: Previous Method}\label{sec:baseline}

The approach we present here is an improvement of the method that we first proposed in \cite{10.1145/3449639.3459308}. We use this as a baseline against which we compare the new method. In its main aspects, the previous method is similar to the new method, with the notable difference that it does not make use of a surrogate model to estimate the distance between two neural network models. As discussed previously, this original method calculates exact distances between each pair of neural networks by first training all the models in the current generation with gradient descent and then getting their error vectors by evaluating them on a validation set. As an additional difference, the previous method would calculate at each iteration an \emph{ensemble selection metric} for each member of the population and then add to the final ensemble the single best-scoring neural network in each generation. The new method maintains an \emph{elite archive}, to which a sample of neural networks from each generation with highest novelty score with respect to this archive, which we call \emph{elite score}, is added at each iteration; novelty scores with respect to the final elite archive are calculated at the end for each of its members and this \emph{ensemble score} is used to select a subset of neural networks which will make up the final ensemble.

We compare the new method proposed in this paper with our previous approach along two main lines. Firstly, we seek to understand whether there is a speedup with the new method as a result of increased efficiency when looking for solutions of similar complexity to those found with the previous method. Secondly, we investigate whether the new method can be used to produce better solutions, i.e. solutions of higher complexity and leading to better final performance. This means that we are interested in investigating whether there is both a quantitative improvement, i.e. being able to do \emph{more of} what could be done with the previous method thanks to a more efficient use of computational resources, and a qualitative improvement, i.e. being able to do \emph{more than} what could be done with the previous method by tackling solutions that were previously unfeasible or intractable.

\section{Experiments}\label{sec:experiments}

This section describes the two sets of experiments carried out for comparing the new NS method, which makes use of a surrogate model to estimate the distance between models as described previously, with the previous method, which instead calculates exact distance values by first training all the models in the population with gradient descent and then determining their error vectors on a validation set. We compare the methods based on resource usage, namely runtime, for similar parameter settings and model complexity, as well as on their ability to scale to larger search spaces and search for more complex models.

\subsection{Test set 1: Resource Usage for Similar Complexity}

In this set of tests, we investigate the total time required to run each of the two methods when they are looking for \emph{solutions of the same complexity} and running for the same number of iterations. We wish to determine the speedup that can be gained with the new method, which makes use of a surrogate model to overcome the need for training all the models in the current generation with gradient descent in order to calculate novelty scores. We run both the new and the previous methods on CIFAR-10 and fix the parameters, as shown in the second column of Table \ref{tab:params}. We conjecture that in these conditions our new method not only results in a speedup due to the use of a Random Forest surrogate model, but also outputs ensembles of similar performance. This is expressed by Hypotheses \ref{hyp:runtime} and \ref{hyp:same_acc}.

\begin{hyp}[Runtime of previous NS method vs new method enhanced with a surrogate model]\label{hyp:runtime}
Enhancing the NS procedure with a Random Forest surrogate model pretrained to estimate the distance between models, and thereby their novelty scores, results in a speedup compared with our previous method, which calculates exact distance values and novelty scores, when constructing ensembles of the same complexity.
\end{hyp}

\begin{hyp}[Performance achieved with the previous NS method vs the new method with a surrogate model]\label{hyp:same_acc}
When looking for solutions of the same complexity, the new NS procedure, using a surrogate model, outputs ensembles which do not perform worse than those constructed by our previous method, even though the new method only \emph{estimates} distance values and novelty scores.
\end{hyp}

\begin{table}[t]
  \caption{Novelty search parameters for both test sets}
  \label{tab:params}
  \begin{tabularx}{\columnwidth}{Y|Y|Y}
    \toprule
    \textbf{Parameter} & \textbf{Test set 1: runtime comparison (both methods)} & \textbf{Test set 2: expanded search space (new method only)}\\
    \midrule
    Iterations & 10 & 100 \\
    \hline
    Final ensemble size & 11 & 40 \\
    \hline
    Population size & 30 & 100 \\
    \hline
    Diversity metric & $\text{cos\_dist}_{i,j}$ & All from Section \ref{sec:metrics} \\
    \hline
    Number of blocks & 2:6 & 2:6 \\
    \hline
    Number of channels in the first convolution & 4:16 & 4:16 \\
    \hline
    Number of channels in residual blocks & 24:32 & 16:64 \\
    \hline
    Dropout probability in residual blocks & 0.1:0.4 & 0.1:0.9 \\
    \hline
    Number of neighbours $K$ & 3 & 15 \\
    \hline
    Size $n_A$ of archive sample & 5 & 10 \\
    \hline
    Size of tournament for selection & 10 & 50 \\
    \bottomrule
    \end{tabularx}
\end{table}

\subsection{Test set 2: Expanding the Search Space}

Using a surrogate model to speed up the procedure has enabled us to both search for solutions of higher complexity and run the NS for longer. In this set of experiments, we apply the new method to three benchmark datasets from the Computer Vision (CV) literature --- CIFAR-10, CIFAR-100, and SVHN --- and test it with all diversity metrics previously defined in Section \ref{sec:metrics}. We also compare the results achieved with the new method to the best results observed with the previous method. The parameters that we use with the new method are shown in the third column of Table \ref{tab:params}; they correspond to the \emph{expanded} search space made possible by the use of a surrogate model. We expect to see further evidence of what we observed in previous work \cite{10.1145/3449639.3459308} regarding \emph{error diversity} metrics, namely that those diversity metrics which focus more closely on the instances where the models make prediction errors lead to higher-performing ensembles. This is expressed by Hypothesis \ref{hyp:error_metrics}. We also expect the new method to lead to higher-performing ensembles than those constructed with the previous method, since the use of a surrogate model makes it feasible to expand the search space and run the NS for longer. This is expressed by Hypothesis \ref{hyp:better_acc}.

\begin{hyp}[Better performance with metrics that focus on error instances]\label{hyp:error_metrics}
In a similar fashion to what we have observed with our previous method, running the NS procedure with the distance metrics that focus more closely on the instances where the models make prediction errors leads to higher-performing ensembles than when more generic diversity metrics are employed.
\end{hyp}

\begin{hyp}[Performance achieved with the previous NS method vs the new method with a surrogate model]\label{hyp:better_acc}
The new NS method enhanced with a surrogate model makes it possible to search a larger space of more complex neural network architectures and, therefore, outputs higher-performing ensembles than the best ones constructed by our previous method.
\end{hyp}

\section{Results and Discussion}\label{sec:results}

In this section, we present the results of the two sets of experiments described in Section \ref{sec:experiments}. We then discuss these results and whether the hypotheses formulated above can be rejected.

\subsection{Hypothesis \ref{hyp:runtime}}

Table \ref{tab:set1_results} shows the median value, calculated after 10 independent runs, of the time required to run both the previous NS method and the new method, which makes use of a surrogate model, with the same parameters. These results show that the new method is about 10 times faster than the original NS method. A Mann-Whitney significance test shows that this difference is significant at the 1\% level. This supports the claim of Hypothesis \ref{hyp:runtime} that enhancing the NS method with a Random Forest surrogate model to estimate the distances between models speeds up the search for diverse models and the construction of a diverse ensemble. For reference, we also report in Table \ref{tab:set1_results} the median time, over 10 runs, required to train a sample of 40 neural network architectures on CIFAR-10, as well as to build a dataset and train the Random Forest surrogate model as per Algorithm \ref{alg:surrogate}. Note that these two runtimes are a \emph{one-off cost} and that, in order to pretrain the surrogate model for our experiments, we have trained a total of 3200 sample architectures by running several processes in parallel on a cluster, each training 40 architectures.

\begin{table}[t]
  \caption{Median results over 10 runs of the previous NS method and the new NS method with a surrogate model on CIFAR-10 (test set 1)}
  \label{tab:set1_results}
  \begin{tabularx}{\columnwidth}{Y|Y}
    \toprule
    \textbf{Runtime of NS} & 48760.5 s\\
    \hline
    \textbf{Runtime of NS with surrogate model} & 4871 s\\
    \hline
    \textbf{Training a sample of architectures} & 28970.5 s\\
    \hline
    \textbf{Building a dataset and training the Random Forest surrogate model} & 18113.5 s\\
    \hline
    \textbf{Accuracy achieved by NS} & 82.245\% \\
    \hline
    \textbf{Accuracy achieved by NS with surrogate model} & 83.885\% \\
    \bottomrule
    \end{tabularx}
\end{table}

\subsection{Hypothesis \ref{hyp:same_acc}}

Table \ref{tab:set1_results} also shows the median accuracy, calculated after 10 independent runs, achieved by ensembles constructed by both the previous NS method and the new method, when these are executed with the same parameters. The results show that the ensembles constructed by the new method do not perform worse than those constructed by the original method, which calculates exact values for the distance metrics and novelty scores. In fact, we observe that the new method leads to slightly better performance. A Mann-Whitney significance test shows that this difference is significant at the 1\% level. This corroborates Hypothesis \ref{hyp:same_acc}, which claims that there is no loss in performance when using the new method and its surrogate estimates. Besides the use of surrogate models, the major difference between the previous and the new method is the way a subset of all the models is selected to be in the final ensemble. As explained before, the previous method applies an \emph{ensemble selection metric} at each iteration of the NS, whereas the new method keeps an \emph{elite archive}, from which the final ensemble is selected in an additional step at the end of the procedure. It seems that the ensemble selection procedure of the new method is the cause behind the better performance achieved by its ensembles.

% \begin{table}[t]
%   \caption{Median accuracy over 10 runs of ensembles constructed by the previous NS method and the new NS method with a surrogate model on CIFAR-10 (test set 1)}
%   \label{tab:accuracy_comp}
%   \begin{tabularx}{\columnwidth}{Y|Y}
%     \toprule
%     \textbf{Accuracy achieved by NS (sec)} & \textbf{Accuracy achieved by NS with surrogate model (sec)}\\
%     \midrule
%     82.245\% & 83.885\% \\
%     \bottomrule
%     \end{tabularx}
% \end{table}

\subsection{Hypothesis \ref{hyp:error_metrics}}

Table \ref{tab:accuracy_new} shows the median accuracy, after 10 runs, of ensembles evolved by the new NS procedure extended with a surrogate model, for all six diversity metrics of Section \ref{sec:metrics} and all three datasets considered. We observe that on CIFAR-10 and SVHN, the metrics $\text{prop}^2_{i,j}$ and $\text{cos\_dist}_{i,j}$ lead to the highest-performing ensembles. Mann-Whitney tests show that the difference to the other metrics is statistically significant. On CIFAR-100, this is observed additionally with the metrics $\text{prop}^{\text{harm}}_{i,j}$ and $\text{dis}_{i,j}$.

The metrics $\text{prop}^2_{i,j}$ and $\text{cos\_dist}_{i,j}$ are the two that focus more closely on the instances where the two models being compared make prediction errors. Additionally, the metric $\text{prop}^{\text{harm}}_{i,j}$ depends on the value of $\text{prop}^2_{i,j}$. These observations back the claim of Hypothesis \ref{hyp:error_metrics} that error diversity metrics lead to better-performing ensembles compared to more generic diversity metrics. This confirms what we observed in our previous work \cite{10.1145/3449639.3459308}.

\begin{table}[t]
  \caption{Median accuracy over 10 runs of ensembles constructed by the new method (test set 2). Best results highlighted. Best results with the original NS shown for comparison}
  \label{tab:accuracy_new}
  \begin{tabularx}{\textwidth}{Y|Y|Y|Y}
    \toprule
    \textbf{Dataset} & \textbf{Diversity metric} & \textbf{Final ensemble accuracy} & \textbf{Best accuracy with original NS (from \cite{10.1145/3449639.3459308})}\\
    \midrule
    \multirow{6}{*}{CIFAR-10} & $\text{prop}^1_{i,j}$ & 67.295\% & \multirow{6}{*}{83.51\%} \\
    \cline{2-3}
     & \textcolor{red}{$\text{prop}^2_{i,j}$} & \textcolor{red}{90.605\%} & \\
    \cline{2-3}
     & $\text{prop}^{\text{harm}}_{i,j}$ & 83.975\% & \\
    \cline{2-3}
     & $\text{dis}_{i,j}$ & 86.28\% & \\
    \cline{2-3}
     & \textcolor{red}{$\text{cos\_dist}_{i,j}$} & \textcolor{red}{90.11\%} & \\
    \cline{2-3}
     & $\text{arch\_dist}_{i,j}$ & 80.4\% & \\
    \midrule
    \multirow{6}{*}{CIFAR-100} & $\text{prop}^1_{i,j}$ & 28.725\% & \multirow{6}{*}{45.42\%} \\
    \cline{2-3}
     & \textcolor{red}{$\text{prop}^2_{i,j}$} & \textcolor{red}{63.05\%} & \\
    \cline{2-3}
     & \textcolor{red}{$\text{prop}^{\text{harm}}_{i,j}$} & \textcolor{red}{63.41\%} & \\
    \cline{2-3}
     & \textcolor{red}{$\text{dis}_{i,j}$} & \textcolor{red}{63.18\%} & \\
    \cline{2-3}
     & \textcolor{red}{$\text{cos\_dist}_{i,j}$} & \textcolor{red}{63.035\%} & \\
    \cline{2-3}
     & $\text{arch\_dist}_{i,j}$ & 49.83\% & \\
    \midrule
    \multirow{6}{*}{SVHN} & $\text{prop}^1_{i,j}$ & 78.825\% & \multirow{6}{*}{91.435\%} \\
    \cline{2-3}
     & \textcolor{red}{$\text{prop}^2_{i,j}$} & \textcolor{red}{94.8\%} & \\
    \cline{2-3}
     & $\text{prop}^{\text{harm}}_{i,j}$ & 89.775\% & \\
    \cline{2-3}
     & $\text{dis}_{i,j}$ & 90.675\% & \\
    \cline{2-3}
     & \textcolor{red}{$\text{cos\_dist}_{i,j}$} & \textcolor{red}{94.79\%} & \\
    \cline{2-3}
     & $\text{arch\_dist}_{i,j}$ & 90.68\% & \\
    \bottomrule
    \end{tabularx}
\end{table}

\subsection{Hypothesis \ref{hyp:better_acc}}

The last column of Table \ref{tab:accuracy_new} shows the best performance achieved by ensembles evolved with our previous NS method. These results show very clearly that the new method constructs higher-performing ensembles than our previous procedure, with the most considerable difference being observed on CIFAR-100 and CIFAR-10. Mann-Whitney tests reveal that, for each dataset, the difference between the best results achieved by the new method and the best achieved by the previous method is indeed statistically significant. This difference results from the fact that the new method, thanks to its use of a surrogate model, is able to \emph{search a wider space of neural network architectures}, even though it runs on \emph{the same bounded resources}. We conclude that this supports Hypothesis \ref{hyp:better_acc}.

% \begin{table}[t]
%   \caption{Best accuracy achieved by the original NS method for each dataset (from \cite{10.1145/3449639.3459308}}
%   \label{tab:accuracy_new}
%   \begin{tabularx}{\textwidth}{Y|Y|Y}
%     \toprule
%     \textbf{Dataset} & \textbf{Diversity metric} & \textbf{Final ensemble accuracy}\\
%     \midrule
%     \multirow{2}{*}{CIFAR-10} & $\text{prop}_{i,j}$ & 83.51\% \\
%     \cline{2-3}
%      & $\text{cos\_dist}_{i,j}$ & 83.29\% \\
%     \midrule
%     \multirow{2}{*}{CIFAR-100} & $\text{prop}_{i,j}$ & 45.42\% \\
%     \cline{2-3}
%      & $\text{cos\_dist}_{i,j}$ & 44.695\% \\
%     \midrule
%     \multirow{2}{*}{SVHN} & $\text{prop}_{i,j}$ & 91.435\% \\
%     \cline{2-3}
%      & $\text{cos\_dist}_{i,j}$ & 91.285\% \\
%     \bottomrule
%     \end{tabularx}
% \end{table}

\section{Conclusions and Future Work}\label{sec:conclusions}

This paper has extended on previous work \cite{10.1145/3449639.3459308}, which proposed an innovative NS method to build behaviourally diverse ensembles of classifiers. The previous method had signposted an innovative way to construct high-performing ensembles by explicitly searching for diversity. However, its application in practice had been hampered by limitations in the amount of available computational resources, since it involved a time-consuming step of training all networks in each generation of the NS with gradient descent. Our new method overcomes this limitation by using a pretrained surrogate model to estimate the distance between neural network architectures, necessary to calculate novelty scores, without the need to train them. In this way, we can obtain an approximate speedup of 10 times w.r.t. the previous method when running them both with the same parameters, \emph{without loss of classification accuracy}. We can also construct better-performing ensembles thanks to the expanded architecture search space facilitated by using a surrogate. We have confirmed previous observations that error diversity metrics lead to better-performing ensembles than more generic metrics.

Our method thus represents an improved paradigm for implementing horizontal scaling of learning algorithms. It makes an explicit search for diversity considerably more tractable than our original approach \emph{for the same bounded resources}. In future work, we will extend the current method by implementing a local competition (LC) variant, so that it is possible to include objectives of accuracy into the NS. This will enable us to further study the relationship between diversity and classification accuracy and to investigate trade-offs between the two. We will also propose more definitions of diversity with new and improved metrics, which will provide more insight into what makes a good diversity metric that fits the task of constructing a diverse high-performing ensemble.

\bibliographystyle{custom.bst}
\bibliography{refs}
\end{document}